\documentclass{article}
\usepackage{spconf,amsmath,graphicx}
\usepackage{siunitx}
\usepackage{wrapfig}
\usepackage{svg}

\newtheorem{problem}{Problem}

\title{Synaptic Partner Assignment Using \\
Attentional Voxel Association Networks}
\name{Author(s) Name(s)}
\address{Author Affiliation(s)}

 \name{Nicholas Turner$^{\star\dagger}$ \qquad Kisuk Lee$^{\dagger}$ \qquad Ran Lu$^{\dagger}$ \qquad Jingpeng Wu$^{\dagger}$ \qquad Dodam Ih$^{\dagger}$ \qquad H. Sebastian Seung$^{\star\dagger}$}

 \address{$^{\star}$ Princeton University - Computer Science Department \\
     $^{\dagger}$ Princeton University - Neuroscience Institute}

\begin{document}
%
\maketitle
\begin{abstract}
Connectomics aims to recover a complete set of synaptic connections within a dataset imaged by volume electron microscopy. Many systems have been proposed for locating synapses, and recent research has included a way to identify the synaptic partners that communicate at a synaptic cleft. We re-frame the problem of identifying synaptic partners as directly generating the mask of the synaptic partners from a given cleft. We train a convolutional network to perform this task. The network takes the local image context and a binary mask representing a single cleft as input. It is trained to produce two binary output masks: one which labels the voxels of the presynaptic partner within the input image, and another similar labeling for the postsynaptic partner. The cleft mask acts as an attentional gating signal for the network. We find that an implementation of this approach performs well on a dataset of mouse somatosensory cortex, and evaluate it as part of a combined system to predict both clefts and connections. 
\end{abstract}
\section{Introduction}
\label{sec:intro}
Connectomics attempts to recover the connectivity of neural circuits from 3D electron microscopic (EM) images of neural tissue \cite{swanson2016cajal}. Recovering the connectome requires segmenting the image into individual neurons, as well as mapping the synaptic connections between neurons. A synapse is a site of apposition (``synaptic cleft'') between the plasma membranes of two neurons, called ``presynaptic'' and ``postsynaptic.'' The interior of the presynaptic cell contains synaptic vesicles (small circular structures) near the cleft, and the membrane on the postsynaptic side of the cleft often features a ``postsynaptic density'' or PSD. The darkness and thickness of the PSD may vary depending on the type of synapse or the quality of the staining. The presynaptic membrane of the cleft is almost always found on a thin kind of neural branch called an axon, typically at a swelling in the axon known as a bouton. Mitochondria are often found in the presynaptic bouton. The postsynaptic membrane of the cleft most commonly belongs to a thick kind of neural branch called a dendrite. Most of the time the postsynaptic membrane is on a thorn-like protrusion from the dendrite called a ``spine.'' The dendritic spine may contain an intracellular organelle called the spine apparatus, which consists mainly of smooth endoplasmic reticulum.

All of the above visual cues are used by human experts to map synapses in EM images. In recent years there have been a number of attempts to automate this process, usually through a two-step approach \cite{dorkenwald2017automated,parag2018detecting,santurkar2017toward,gray2015automated,kreshuk2015talking} though there are alternatives \cite{buhmann2018synaptic,staffler2017synem}. The first step is detection and localization of synaptic clefts. The second step---the focus of the present work---assigns to each synaptic cleft a presynaptic cell and a postsynaptic cell. We refer to this second step as synaptic partner assignment.

Almost all machine learning approaches have formulated the problem as classification of candidate synaptic partners:
\begin{problem}[Candidate partner classification] Given an EM image and an ordered pair of objects as binary masks, determine whether or not the first object is presynaptic cell and the second object is postsynaptic for the same synaptic cleft.
\end{problem}
Machine learning classifiers have been trained for this task based on hand-designed features \cite{dorkenwald2017automated,santurkar2017toward,kreshuk2015talking,huang2018fully}. More recently, convolutional networks were trained for this task by Parag et al. \cite{parag2018detecting}.

\begin{figure}[htb]
\begin{minipage}[b]{1.0\linewidth}
  \centering
  \centerline{\includegraphics[width=8.5cm]{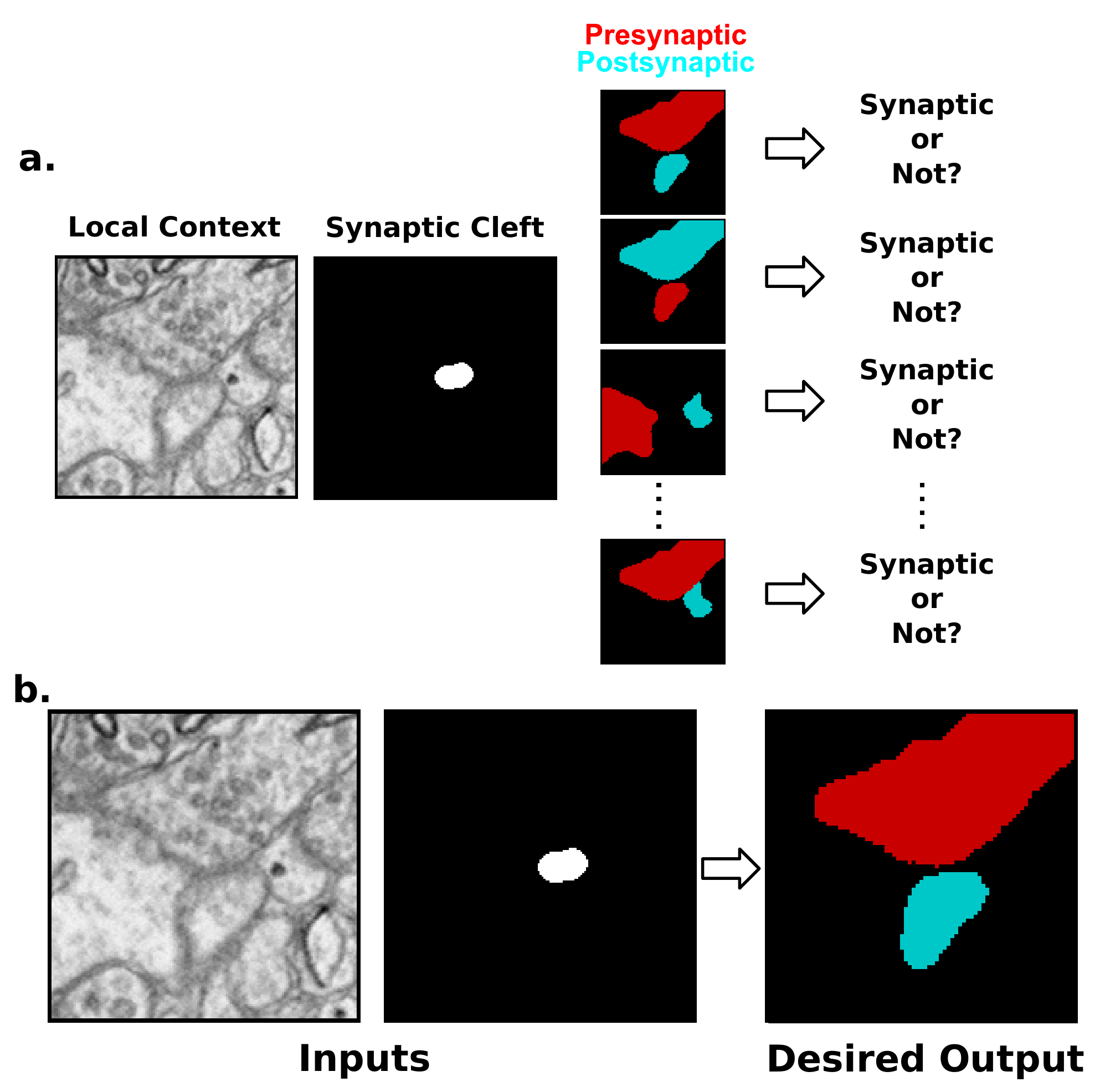}}
\end{minipage}
\caption{(a) The candidate partner classification task. Several pairs of candidate segment pairs are classified as synaptic or non-synaptic. (b) The voxel association task. Presynaptic and postsynaptic partners are generated from the synaptic cleft object that connects them. These masks are matched with existing segments to assign synaptic partner IDs. Although depicted using 2D images, each process occurs in 3D.}
\label{fig:approach}
\end{figure}

In the present work, we formulate the learned task as follows: 
\begin{problem}[Voxel association task]
Given an EM image and a synaptic cleft as a binary mask, generate the presynaptic and postsynaptic objects as binary masks. 
\end{problem}
We call this the voxel association task, because the output images encode object relations as associations between voxels. We train a convolutional network to perform the voxel association task for the Kasthuri et al. \cite{kasthuri2015saturated} dataset of images from mouse somatosensory cortex. The network achieves very low error rates in multiple evaluations, suggesting that our approach could be sufficiently accurate for practical applications in connectomics of the mammalian cortex. Synaptic partner assignment can be more challenging when the PSD is unclear, in organisms with more than two synaptic partners per synapse, and/or images with inferior staining quality. A preliminary application of our approach in this more challenging setting is given in the Supplementary Material, but a full account is postponed to a future paper.

The convolutional net trained by Parag et al. \cite{parag2018detecting} to perform candidate partner classification for the mouse somatosensory cortex (called the ``proximity pruner'' below) also achieves low error on the test set. The difference is not statistically significant due to limited test set size. Therefore, a choice between the two approaches at the present time should be based on considerations other than accuracy. In Problem 2, the convolutional net is applied once per synaptic cleft. In Problem 1, the convolutional net is applied multiple times per cleft (Figure \ref{fig:approach}). On the other hand, a simpler convolutional net may suffice for Problem 1, because it is likely to be a simpler task since the output is a single binary variable. The balance of the preceding considerations will determine which approach requires more computational resources. The voxel association task is potentially simpler to apply, because it avoids the complexity of the candidate suggestion process, which includes a number of manually set thresholds.





\subsection{Related Work}

Mapping synapses is related to visual scene analysis, the computation of relationships between objects in an image \cite{dai2017detecting, xu2017scene}. Our work differs in its focus on a small set of relationships (presynaptic and postsynaptic). Bounding boxes, commonly computed for scene analysis and instance segmentation \cite{ren2015faster}, could also be used as an intermediate or output representation for synaptic cleft detection, because clefts are usually fairly well-separated from each other. However, bounding boxes are not convenient for neurons, because neuronal branches are highly entangled.


Our voxel association task is also similar to the object mask extension task in computer vision \cite{romeraparedes2015recurrent, ren1605end} and connectomics \cite{januszewski2016flood,meirovitch2016multi}, which iteratively grows morphological segments from seed voxels. The voxel association task relies on a single forward pass to sufficiently identify the nearby voxels of interest, yet an iterative procedure could also be useful for difficult examples, or for synapses that are too large to fit in a single context window.

\section{Attentional Voxel Association Networks}





We train a convolutional net to perform this task using supervised learning. We call the trained network an Attentional Voxel Association Network (AVAN). 

\subsection{Training \& Inference}

The training procedure requires an EM dataset with (1) a morphological segmentation, (2) a synaptic cleft segmentation, and (3) a set of ID pairings which match each cleft segment ID with its presynaptic and postsynaptic IDs within the morphological segments. We take the voxels that are part of some synaptic cleft within a training volume, and form training samples from these locations by centering a patch of local context on each location. We create the cleft mask input from the synaptic cleft at the sample location. We generate target output by creating a mask of the segments with the desired pre- or postsynaptic segment ID within the local context region (one volume of output for a presynaptic ID, and another for a postsynaptic ID). 

Each network in this work was implemented using PyTorch \cite{paszke2017automatic}. Our association network takes a $2 \times 80 \times 80 \times 18$ window as input (concatenating image context and seed object), and produces a volume of the same size as output. We used a modified version of the Residual Symmetric U-Net architecture \cite{lee2017superhuman}, where the upsampling operation is replaced by a ``resize convolution'' \cite{odena2016deconvolution}, and one level is removed from the downsampling hierarchy. All networks were trained using a cross-entropy loss function and an Adam optimizer \cite{kingma2014adam} using a manual learning rate schedule.

During inference, the network is provided with patches of input centered around the centroid of each synaptic cleft. The presynaptic segment is predicted as the segment whose voxels have the highest average output over the presynaptic volume output patch, and a similar judgment is performed for the postsynaptic segment. These judgments are restricted to segments which overlap with a dilated version of the target cleft. We used a human annotated cleft segmentation to train the association task, although the training is agnostic to the source of this segmentation.

\subsection{Polyadic Synapses}

Synapses of other organisms can have multiple pre- or postsynaptic partners for each synaptic cleft \cite{buhmann2018synaptic, huang2018fully}. This approach can be extended to handle multiple segments by selecting all segments with average output over a tuned threshold value. We've omitted this extension here for clarity, yet preliminary experiments have shown this approach to be competitive with other published methods (Supplementary Materials).

\section{Evaluation}

We evaluated our basic approach using the Kasthuri et al. dataset \cite{kasthuri2015saturated}, downsampled to a voxel resolution of $\SI{12}{\nano\metre} \times \SI{12}{\nano\metre} \times \SI{30}{\nano\metre}$. This dataset includes two small labeled volumes of voxel dimensions $512 \times 512 \times 100$ and $512 \times 512 \times 256$. An augmented version of these cutouts has been used for comparison of similar systems \cite{parag2018detecting}. We split the labeled cutouts, using the first volume and the top 100 slices of the second as a training set, the middle 56 slices of the second as a validation set, and the bottom 100 as a test set. We labeled synaptic clefts and their partners as required for this task.

We evaluated the trained association network in three ways: (1) assigning human annotated cleft segments, (2) combining it with cleft detection systems, and (3) applying it to an automated reconstruction of the larger image volume.

\subsection{Synapse Assignment}
For comparison, we implemented the candidate classification approach of Parag et al. \cite{parag2018detecting}. In their work, a heuristic suggests candidate partners for each synapse based on overlap of predicted sites with nearby cell segments along with other variables. A ``proximity pruner'' model then classifies each potential pair of candidate partners, using 4 inputs: local image context, the output of their synapse detector, and the pair of candidate segment masks.

We chose this work due to its recent results within this dataset. We transformed our synaptic cleft labels to signed proximity, and trained a pruning network on a set of candidate segment pairs, formed by taking distinct pairs which overlapped with a dilated version of each cleft. We also attempted another candidate classification approach for this task, by training a similar classifier network that takes singular cleft segment masks as input instead of signed proximity (a ``mask pruner'' network).

\begin{table}
\centering
\caption{Assignment Test Results (163 synapses).}\label{tab:assign_res}
\begin{tabular}{|c|c|} \hline  
\bfseries{Assignment Method} & \bfseries{Accuracy} (\%) \\ \hline
Mask Pruner & 96.9 \\ \hline
Proximity Pruner \cite{parag2018detecting} & 97.5 \\ \hline
AVAN & 100 \\ \hline
\end{tabular}
\end{table}

During this evaluation, we modified the inference procedure for the pruning networks. Instead of a threshold scheme, each approach took the segment pair with highest output for each ground truth cleft segment as its prediction. This allows all methods to take the identity and number of true synapses as given, and also allows us to use accuracy metrics to compare them.

Each of these three methods performed very well and within error of this small dataset (Table \ref{tab:assign_res}). Notably, our modified classification approach rivaled the performance of the original version, and both classification networks performed slightly worse than the association network. This suggests little overall benefit to assignment from adding signed information, and competitive performance for each approach.

\subsection{Combined System}

Next, we trained models to predict the locations of synaptic clefts, and then combined these predictions with each form of partner assignment. This step explored whether imperfectly predicted seed objects would affect assignment. We also included another training representation to produce cleft masks in an attempt to reproduce its effects on performance \cite{heinrich2018synaptic}.
 
We've found that na\"ive methods of synapse annotation can be inaccurate, as attentional gaps can miss examples \cite{plaza2014annotating}. Our cleft ground truth was initially labeled by humans, and then ``bootstrapped'' twice. Each round of bootstrapping involved reviewing the errors of each cleft prediction method after a round of training, and including valid examples into the labels. This increased the number of examples in the training, validation, and test sets by 18, 17, and 15 respectively. All cleft networks were fine-tuned with the updated ground truth after the final additions.

To evaluate each combined system, we performed a grid search over its hyperparameters, optimizing the F-1 score of predicted graph edges. A correct edge was defined as a segment (or pair of segments) that maximally overlaps with a cleft label and is assigned to the correct partners. Each cleft output was generated by a trained model with the same Residual Symmetric U-Net architecture.
\begin{figure}[htb]

\begin{minipage}[b]{1.0\linewidth}
  \centering
  \centerline{\includegraphics[width=\textwidth]{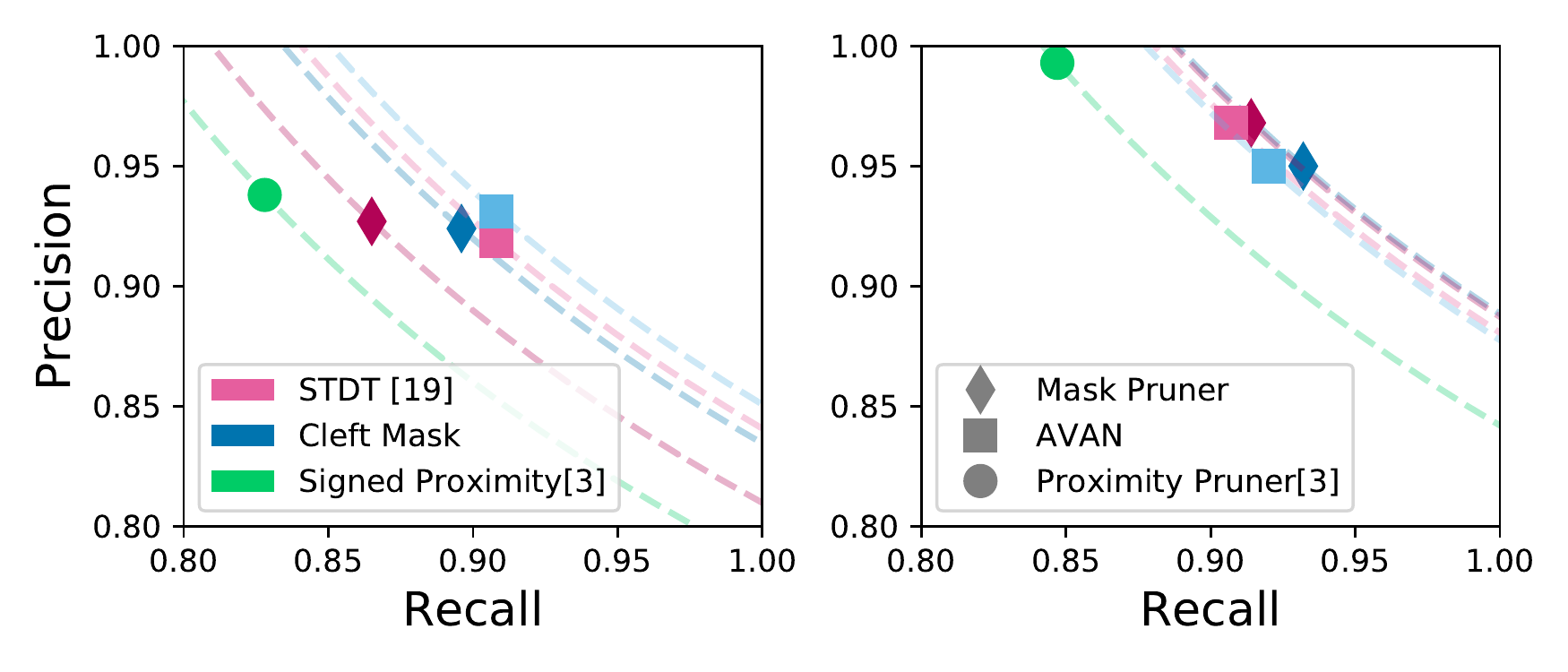}}
\end{minipage}
\caption{Combined System Evaluation (163 synapses). \emph{(left)} Combined system performance optimized for F-1 score. \emph{(right)} Cleft detection performance optimized for combined system F-1 score. Color indicates cleft detection target representation. Shape indicates partner assignment method. Dashed lines show level sets of F-1 score at the value for each system.}
\label{fig:combined_res}
\end{figure}
 
The performance of the combined systems showed more variability than the assignment task alone (Figure \ref{fig:combined_res}). By inspecting the cleft detection scores at the final hyperparameter settings and performing an error analysis, we found that the differences between methods were mostly due to cleft detection instead of assignment. While the STDT representation and the cleft mask had similar performance, signed proximity output produced more false negatives. 

\subsection{Large-Scale Evaluation}

\begin{figure}[htb]
\begin{minipage}[b]{1.0\linewidth}
  \centering
  \centerline{\includegraphics[width=8.5cm]{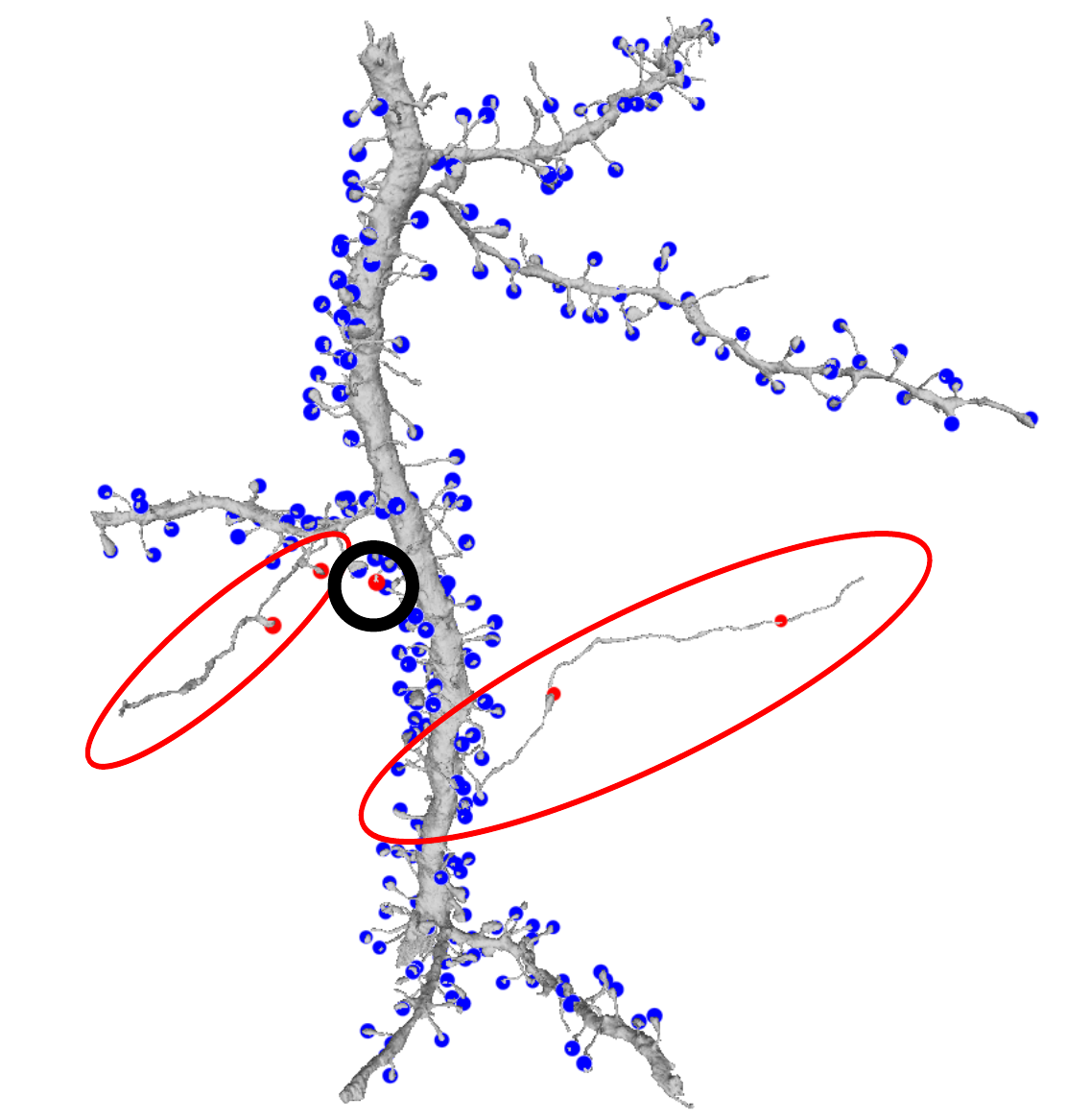}}
\end{minipage}
\caption{Large-Scale Evaluation. (a) Example reconstructed dendrite with its assigned synapses. Dendrites should only have postsynaptic terminals. Red/blue dots: pre/postsynaptic terminals. Presynaptic terminals target axonal segments incorrectly merged to the dendrite by the segmentation algorithm (red ovals) and one error (black circle) from a false positive cleft.}
\label{fig:large_scale}
\end{figure}

We observed high accuracy of synaptic partner assignment methods in small cutouts. Attempting to mine for harder examples, we further tested our assignment method by applying it to a larger volume. We performed an automated reconstruction of the entire Kasthuri et al. \cite{kasthuri2015saturated} dataset (Fig \ref{fig:large_scale}), predicted synaptic clefts using a cleft mask network, and selected 1000 predicted synapses at random from this result. We applied the AVAN at each selected synapse, as well as an early version of our proximity prediction network and the fully trained proximity pruner.

Proofreading the disagreements to remove false positive clefts left 81 locations where the two methods initially predicted different partners. We annotated each of these examples, as well as an equal number of false negative clefts to ensure that our measurement wasn't biased towards locations that the cleft detector is likely to find. We also transformed each cleft to signed proximity in order to fairly apply the proximity pruning approach. Out of this new set of 162 synapses we found 3 mistakes from the association network, and 4 from the proximity pruner.

\section{Discussion}

We've presented a method for synapse assignment which performs competitively within multiple benchmarks. This serves to establish a proof of principle for our approach. Notably, this method produces reliable results without signed input, hand-designed features, or multi-step candidate suggestion procedures.

Our comparisons are specific to the Kasthuri et al. \cite{kasthuri2015saturated} dataset, yet our findings suggest that the dominant source of error is cleft detection for well-stained diadic (two-partner) synapses. Preliminary experiments have shown that our approach also works well in polyadic (multi-partner) systems (Supplemental Materials), yet future work will address general challenges.


\end{document}